# A 3D-2D-convolutional Neural Network Model for Hyperspectral Image Classification

Jiaxin Cao[1] and Xiaoyan Li[1]

[1] School of Computer Science, China University of Geosciences, Wuhan 430074, China

**Abstract:** In recent years, with the emergence of new technologies for deep learning, deep learning is widely used for hyperspectral image classification. convolutional neural network is one of the most frequently used deep learning based methods for visual data processing. The use of CNN for hyperspectral image classification is also visible in recent works. However, the classification effect is not satisfactory when limited training samples are available. Focused on "small sample" hyperspectral classification, we proposed a novel 3D-2D-convolutional neural network model named SE-HybridSN. In SE-HybridSN model, a dense block was used to reuse shallow features and aimed at better exploiting hierarchical spatial–spectral features. Subsequent depth separable convolutional layers were used to discriminate the spatial information. Further refinement of spatial–spectral features was realized by the channel attention method, which were performed behind every 3D convolutional layer and every 2D convolutional layer. Experiment results indicate that our proposed model can learn more discriminative spatial–spectral features using very few training data. In Indian Pines, Salinas and the University of Pavia, SE-HybridSN using only 5%, 1% and 1%labeled data for training. A very satisfactory performance is obtained using the proposed SE-HybridSN.

**Keywords:** Hyperspectral Image Classification, Convolutional Neural Networks, Deep Learning, Spatial–Spectral, Channel Attention

## 1 Introduction

Hyperspectral image classification is the hotspot in remote sensing image interpretation and is of great difficulty. Its purpose is to assign an accurate label to each pixel in the image and then divide the image into areas with different ground object semantic identification [1]. Currently, the convolutional neural network has been successfully applied to the tasks of hyperspectral image classification [3-5]. In hyperspectral image (HSI) classification, the convolutional neural network acts as an "information distiller", gradually extracting high-level abstract semantic features with the deepening of the network. In this process, the hyperspectral images with a huge amount of data are transformed, the irrelevant information is filtered out, and the useful information is enlarged and refined [7]. Prior to deep learning methods, traditional methods mostly used a linear discriminant analysis [8], such as the principal component analysis [9] and independent component analysis [10], to extract features. Additionally, they used a shallow classifier [11–13] to complete classification. These methods rely on manual designed features. For complex and diverse hyperspectral data, it is difficult to find a universal feature extraction method using such a route. Convolution neural network, which can learn features from HSI autonomously, provides a good solution for feature extraction. The HSI classification models based on 1D-CNN [14] or 2D-CNN [15] can achieve considerable classification results by automatically extracting features from hyperspectral images, but along with a degree of spatial or spectral information loss.

Chen et al. [18] constructed a 3D-CNN model composed of 3D convolutional layers and 3D pooling layers, improving classification performance by means of deep exploration into spatial–spectral features. Deeper networks enable deeper and more robust features and the network structure needs careful designing to pretend the greatly rising of the parameters amount. Lee et al. [19] made good use of residual connection in the spectral feature learning and built a deeper network (Res-2D-CNN) by which deeper and more abstract features could be extracted. Liu et al. [31] introduce residual connections to 3D-CNN and built Res-3D-CNN, which is aimed at enhancing spatial–spectral feature learning. Zhong et al. [20] focused on the raw hyperspectral data without dimensionality reduction and built SSRN (spectral–spatial residual network). They introduced residual connection into the whole network and separate deep feature learning procedure into independent spatial feature learning and spectral feature learning. More discriminative features were learned by SSRN and the separated feature learning pattern has a significant impact on subsequent hyperspectral classification research. Recently, dense connections have attracted more attention from hyperspectral researchers [32]. Dense connection reduces the network parameters through a small convolution kernel number, and realizes efficient feature reuse through feature map concatenation, both of which alleviates the problem of model overfitting.

Hu et al. [20] proposed squeeze-and-excitation networks and introduced the attention mechanism to the image classification network, winning the champion of 2017 ImageNet Large Scale Visual Recognition Competition. Recently, the attention mechanism [21] has been applied to the construction of HSI classification models. The attention mechanism is a resource allocation scheme, through which limited computing resources can be used to process more important information. Therefore, the attention mechanism module can effectively enhance the expression ability of the model without excessively increasing complexity. Wang et al. [22] constructed a spatial–spectral squeeze-and-excitation (SSSE) module to automatically learn the weight of different spectral and different neighborhood pixels to emphasize the meaningful features and suppress unnecessary ones so that the classification accuracy is improved effectively. Li et al. [23] added an attention module (Squeeze-and-Excitation block) respectively after the dense connection module used for shallow and middle feature extraction to emphasize effective features in the spectral bands, and then feed it to further deep feature extraction. The attention mechanism in the HSI classification model is used for finding more discriminative feature patterns in spectral or spatial dimension

Based on 3D-2D-CNN and the densely connected module, SE-HybridSN realized a more efficient feature reuse and feature fusion. Moreover, the attention mechanism was introduced to the 3D Based on 3D-2D-CNN and the densely connected module, AD-HybridSN realized a more efficient feature reuse and feature fusion. Moreover, the attention mechanism was introduced to the 3D and spatial features in a targeted refinement circumstance. With fewer parameters, SE-HybridSN achieves better classification performance in the Indian Pines, Salinas and University of Pavia datasets.

## 2    Method[1]

### 2.1    SE-HybridSN Model

Hyperspectral image classification is to assign a specific label to every pixel in hyperspectral images. The convolutional neural networks based hyperspectral classification models take small image patches as the input. Every hyperspectral image patch was composed of the spectral vectors within a predefined range and its land-use type was determined by its center pixel. The hyperspectral patch can be denoted as $P_{L \times W \times S}$, where $L \times W$ represents the spatial dimension and S represents the number of spectral bands. In our proposed model, the input data was processed by principal components analysis (PCA) in the spectral dimension, which greatly reduced the redundancy within hyperspectral data. Figure 1 shows the network structure of the proposed SE-HybridSN. SE-HybridSN is based on the 3D-2D-CNN feature extraction pattern and is composed of 6 convolutional layers. We introduced the channel attention method after every 3D convolutional layer to refine the extracted spatial–spectral features.

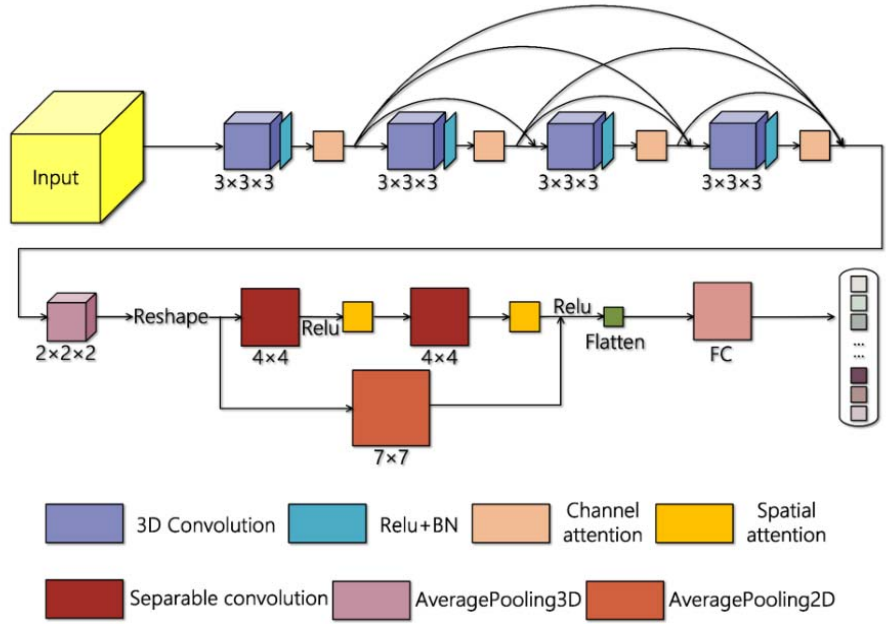

**Fig. 1.**

## 2.2 Convolutional Layers Used in Model

In 2D-CNN, the input data are convolved with 2D kernels. The convolution happens by computing the sum of the dot product between input data and kernel. The kernel is strided over the input data to cover full spatial dimension. The convolved features are passed through the activation function to introduce the non-linearity in the model. In 2D convolution, the activation value at spatial position $(x, y)$ in the $j^{th}$ feature map of the $i^{th}$ layer, denoted as $v_{i,j}^{x,y}$, is generated using the following equation,

$$v_{i,j}^{x,y} = \varphi(b_{i,j} + \sum_{\tau=1}^{d_{l-1}} \sum_{\rho=-\gamma}^{\gamma} \sum_{\sigma=-\delta}^{\delta} w_{i,j,\tau}^{\sigma,\rho} \times v_{i-1,\tau}^{x+\sigma,y+\rho}) \tag{1}$$

The 3D convolution is done by convolving a 3D kernel with the 3D-data. In the proposed model for HSI data, the feature maps of convolution layer are generated using the 3D kernel over multiple contiguous bands in the input layer; this captures the spectral information. In 3D convolution, the activation value at spatial position $(x, y, z)$ in the $j^{th}$ feature map of the $i^{th}$ layer, denoted as $v_{i,j}^{x,y,z}$, is generated as follows,

$$v_{i,j}^{x,y,z} = \varphi(b_{i,j} + \sum_{\tau=1}^{d_{l-1}} \sum_{\lambda=-\eta}^{\eta} \sum_{\rho=-\gamma}^{\gamma} \sum_{\sigma=-\delta}^{\delta} W_{i,j,\tau}^{\sigma,\rho,\lambda} \times v_{i-1,\tau}^{x+\sigma,y+\rho,z+\lambda}) \tag{2}$$

where $2\eta + 1$ is the depth of kernel along spectral dimension and other parameters are the same as in (Eqn. 1).

## 2.3 Attention Mechanism

Currently, the attention mechanism has been successfully applied to the area of computer vision based on convolutional neural networks. The attention mechanism can be used to readjust feature maps generated by some layers of a neural network, which make it able to detect specific channel or spatial feature. The attention mechanism can be roughly divided into spatial attention and channel attention. In our proposed model, channel attention was introduced to the refactor and refines the spatial–spectral features extract by every convolutional layer in the dense block.

**Channel Attention Module**

As mentioned above, the feature map extracted by a single 3D convolution kernel is modeled as a 3D cube, which can learn detailed features and correlation information across spectral bands of hyperspectral data to some extent. Take the 3 × 3 × 3 convolu-tional kernel as an example, the same parameters are used for single channel 3D hyperspectral data during which each convolution operation covers three spectral bands. As the band span of the spectral features characterized by a single convolutional layer is fixed, the spectral feature mining has been limited to some extent. Therefore, in order to further refine the extracted spatial–spectral features, feature maps of all channels were concatenated in the spectral dimension to form a 3D tensor. The reshaped 3D tensor had a large channel number, which was equal to the original channel number times the original spectral band number. Then channel attention was introduced to assign a specific weight for each channel.

Figure 2 is a schematic diagram of the channel attention mechanism used in this article. Let the dimension of the feature map generated by 3D convolutional layers be $B \times L \times W \times C \times N$, where B represents batch size, $L \times W$ represents the spatial dimension of the feature map, C represents the spectral dimension and N represents the number of convolution kernels. In our proposed method, the 5D feature map will be reshaped to be a 4D tensor and its dimension will be $B \times L \times W \times (CN)$, where CN is the new channel number.

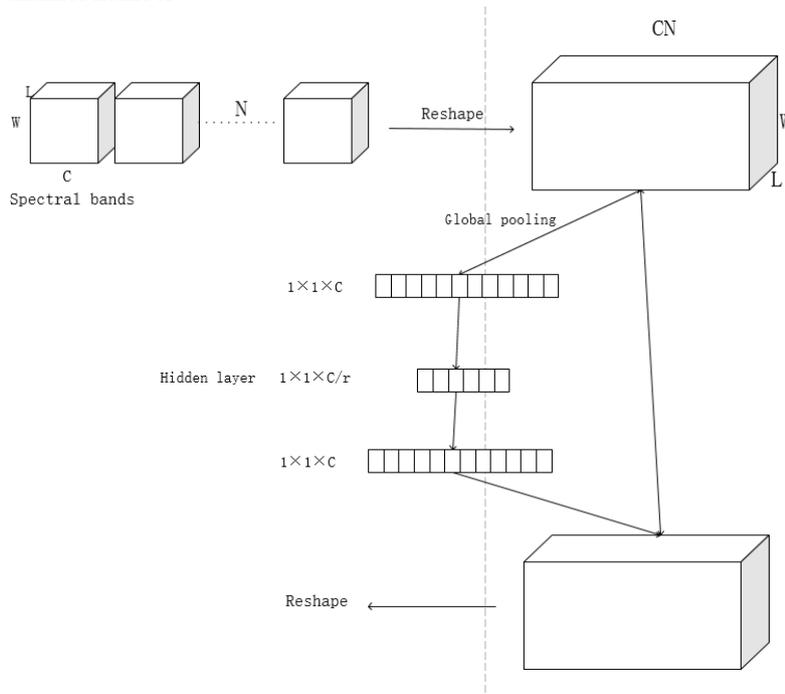

**Fig. 2.** The overall architecture of the channel attention mechanism used in SE-HybridSN

## 3    Experiments

### 3.1    Dataset

We have used three publicly available hyperspectral image datasets, namely Indian Pines, University of Pavia and Salinas Scene. The Indian Pines (IP) dataset has images with 145 × 145 spatial dimension and 224 spectral bands in the wavelength range of 400 to 2500 nm, out of which 24 spectral bands covering the region of water absorption have been discarded. The ground truth available is designated into 16 classes of vegetation. The University of Pavia (UP) dataset consists of 610×340 spatial dimension pixels with 103 spectral bands in the wavelength range of 430 to 860 nm. The ground truth is divided into 9 urban land-cover classes. The Salinas Scene (SA) dataset contains the images with 512×217 spatial dimension and 224 spectral bands in the wavelength range of 360 to 2500 nm. The 20 water absorbing spectral bands have been discarded. In total 16 classes are present in this dataset. Figures 4–6 show the distribution of each ground objects on the Indian Pines, University of Pavia and Salinas.

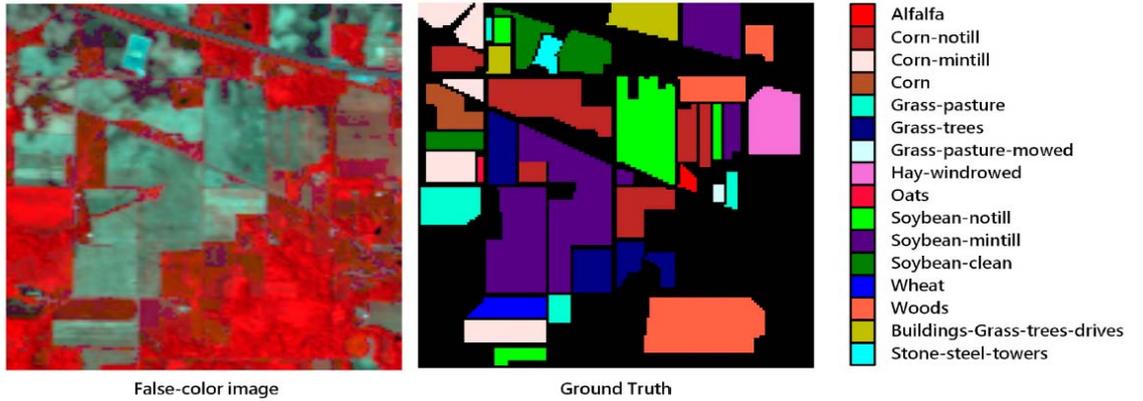

| | |
|---|---|
| | Alfalfa |
| | Corn-notill |
| | Corn-mintill |
| | Corn |
| | Grass-pasture |
| | Grass-trees |
| | Grass-pasture-mowed |
| | Hay-windrowed |
| | Oats |
| | Soybean-notill |
| | Soybean-mintill |
| | Soybean-clean |
| | Wheat |
| | Woods |
| | Buildings-Grass-trees-drives |
| | Stone-steel-towers |

False-color image            Ground Truth

**Fig. 3.** False-color image and color coding for Indian Pines

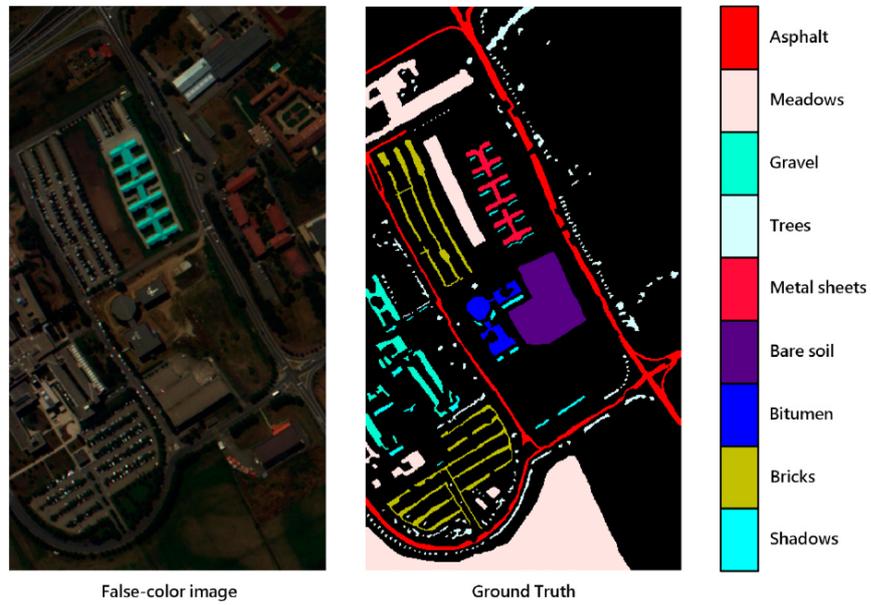

False-color image     Ground Truth

| | |
|---|---|
| ■ | Asphalt |
| ■ | Meadows |
| ■ | Gravel |
| ■ | Trees |
| ■ | Metal sheets |
| ■ | Bare soil |
| ■ | Bitumen |
| ■ | Bricks |
| ■ | Shadows |

**Fig. 4.** False-color image and color coding for the University of Pavia

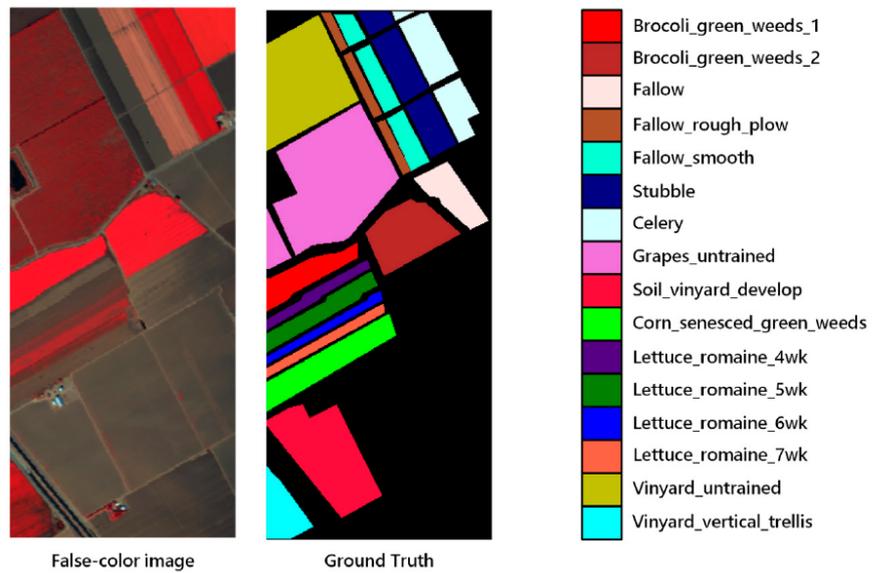

False-color image     Ground Truth

| | |
|---|---|
| ■ | Brocoli_green_weeds_1 |
| ■ | Brocoli_green_weeds_2 |
| ■ | Fallow |
| ■ | Fallow_rough_plow |
| ■ | Fallow_smooth |
| ■ | Stubble |
| ■ | Celery |
| ■ | Grapes_untrained |
| ■ | Soil_vinyard_develop |
| ■ | Corn_senesced_green_weeds |
| ■ | Lettuce_romaine_4wk |
| ■ | Lettuce_romaine_5wk |
| ■ | Lettuce_romaine_6wk |
| ■ | Lettuce_romaine_7wk |
| ■ | Vinyard_untrained |
| ■ | Vinyard_vertical_trellis |

**Fig. 5.** False-color image and color coding for Salinas

In our experiments, for each image we divided all the pixels into three parts: training set, test set and validation set. The proportion of the training set and validation set of the Indian Pines, University of Pavia and Salinas datasets was 5%, 1% and 1% respectively and the remaining pixels served as a test set. The sample distribution of the three datasets for each class of ground object is shown in Tables 1–3.

**Table 1.** Number of training, validation and testing samples of the Indian Pines

| Number | Class | Training | Validation | Testing | Total |
|---|---|---|---|---|---|
| 1 | Alfalfa | 3 | 2 | 41 | 46 |
| 2 | Corn-notill | 72 | 71 | 1285 | 1428 |
| 3 | Corn-min | 41 | 42 | 747 | 830 |
| 4 | Corn | 12 | 12 | 213 | 237 |
| 5 | Grass/Pasture | 24 | 24 | 435 | 483 |
| 6 | Grass/Trees | 37 | 36 | 657 | 730 |
| 7 | Grass/Pasture-mowed | 2 | 1 | 25 | 28 |
| 8 | Hay-windrowed | 24 | 24 | 430 | 478 |
| 9 | Oats | 1 | 1 | 18 | 20 |
| 10 | Soybean-notill | 48 | 49 | 875 | 972 |
| 11 | Soybean-mintill | 122 | 123 | 2210 | 2455 |
| 12 | Soybean-clean | 30 | 29 | 534 | 593 |
| 13 | Wheat | 10 | 10 | 185 | 205 |
| 14 | Woods | 63 | 63 | 1139 | 1265 |
| 15 | Building-Grass-Trees-Drives | 20 | 19 | 347 | 386 |
| 16 | Stone-steel Towers | 4 | 5 | 84 | 93 |
| | Total | 512 | 512 | 9225 | 10249 |

**Table 2.** Number of training, validation and testing samples of the University of Pavia

| Number | Class | Training | Validation | Testing | Total |
|---|---|---|---|---|---|
| 1 | Asphalt | 66 | 66 | 6499 | 6631 |
| 2 | Meadows | 186 | 186 | 18277 | 18649 |
| 3 | Gravel | 21 | 21 | 2057 | 2099 |
| 4 | Trees | 30 | 31 | 3003 | 3064 |
| 5 | Metal sheets | 14 | 13 | 1318 | 1354 |
| 6 | Bare soil | 50 | 50 | 4929 | 5029 |
| 7 | Bitumen | 13 | 14 | 1303 | 1330 |
| 8 | Bricks | 37 | 37 | 3608 | 3682 |

| 9 | Shadows | 10 | 9 | 928 | 947 |
| | Total | 427 | 427 | 41922 | 42776 |

**Table 3.** Number of training, validation and testing samples of the Salinas

| Num-ber | Class | Training | Validation | Testing | Total |
|---|---|---|---|---|---|
| 1 | Brocoli_green_weeds_1 | 20 | 20 | 1969 | 2009 |
| 2 | Brocoli_green_weeds_2 | 37 | 37 | 3625 | 3726 |
| 3 | Fallow | 20 | 20 | 1936 | 1976 |
| 4 | Fallow_rough_plow | 14 | 14 | 1366 | 1394 |
| 5 | Fallow_smooth | 27 | 27 | 2624 | 2678 |
| 6 | Stubble | 40 | 39 | 3880 | 3959 |
| 7 | Celery | 36 | 36 | 3507 | 3579 |
| 8 | Grapes_untrained | 112 | 113 | 11046 | 11271 |
| 9 | Soil_vinyard_develop | 62 | 62 | 6079 | 6203 |
| 10 | Corn_senesced_green_weeds | 33 | 33 | 3212 | 3278 |
| 11 | Lettuce_romaine_4wk | 11 | 10 | 1047 | 1068 |
| 12 | Lettuce_romaine_5wk | 19 | 20 | 1888 | 1927 |
| 13 | Lettuce_romaine_6wk | 9 | 9 | 898 | 916 |
| 14 | Lettuce_romaine_7wk | 10 | 11 | 1049 | 1070 |
| 15 | Vinyard_untrained | 72 | 73 | 7213 | 7268 |
| 16 | Vinyard_vertical_trellis | 18 | 18 | 1771 | 1807 |
| | Total | 541 | 541 | 53047 | 54129 |

### 3.2 Experimental Results

Three indexes were used to measure the accuracy of models, namely, overall accuracy (OA), average accuracy (AA) and Kappa coefficient (Kappa). OA represents the proportion of the number of samples that were correctly classified by the model. AA stands for the average precision of all land objects. KAPPA is an accuracy measure based on the confusion-matrix, which represents the percentage of errors reduced by classification versus a completely random classification.

In order to avoid fluctuations caused by accidental factors as far as possible, we conducted 20 consecutive experiments. Tables 4–6 show the average indices and standard deviation of each model on three datasets. Figures 7–9 show the false-color map, the ground truths and the classification results of each model for three datasets. We can tell by the data and predicted maps that the classification result of SE-HybridSN was more detailed and accurate in Indian Pines, Salinas and University of Pavia. Among the contrast models, the OA of 2D-CNN on the three datasets were lower than the other con-trast models, indicating that the 2D-CNN model was not suitable for small sample hyperspectral classification. Secondly, the classification result of Res-3D-CNN was higher than that of Res-2D-CNN, indicating that the 3D-CNN model could explore spatial–spectral features of training samples more effectively. R-HybridSN was superior to the HybridSN in Indian Pines and University of Pavia, and the two models had a higher classification accuracy than 3D-CNN, to a certain extent, it proved that, compared with the model that used the 3D convolution kernel or 2D convolution kernel alone, the 3D-2D-CNN model was more suitable for the classification under the condition of small samples. Among the three 3D-2D-CNN models, SE-HybridSN achieved the highest classification accuracies in three datasets. For example, the OA of SE-HybridSN was 2.63% and 3.66% higher than HybridSN and SSRN in Indian Pines.

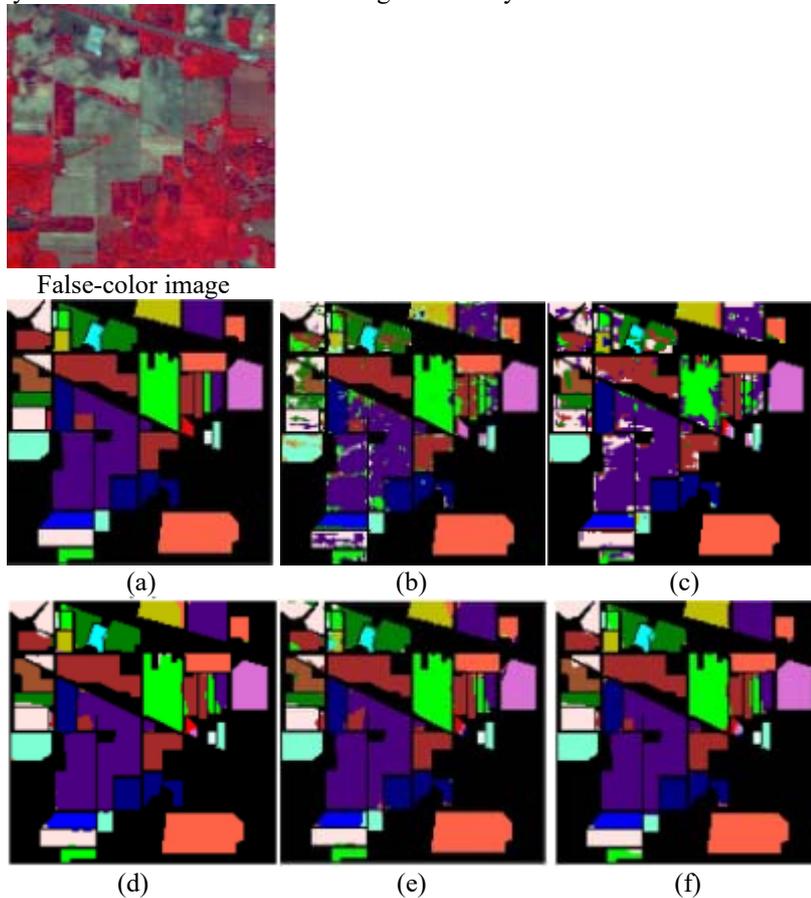

**Fig. 6.** The classification maps of Indian Pines. (**a**) Ground truth. (**b–f**) Predicted classification maps for 2D-CNN, 3D-CNN, SSRN, HybridSN and SE-HybridSN respectively.

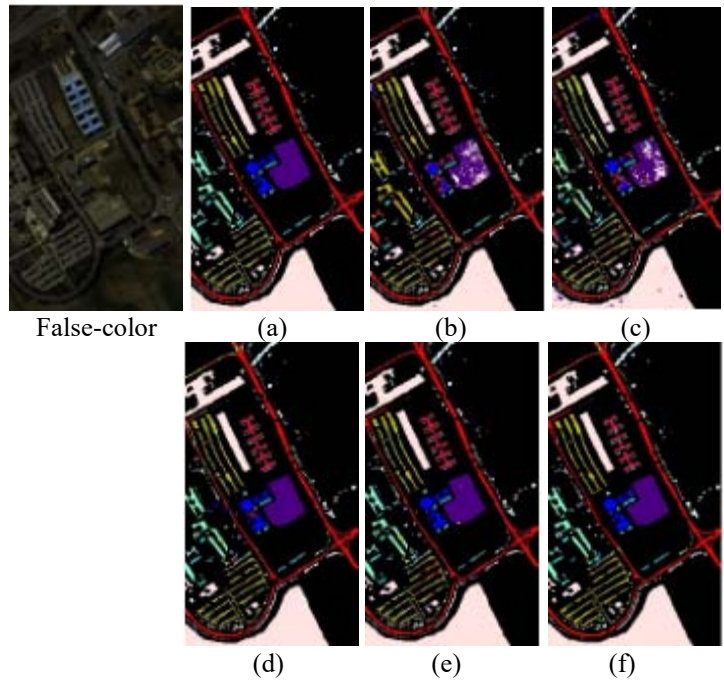

**Fig. 7.** The classification maps of the University of Pavia. (**a**) Ground truth. (**b–f**) Predicted classification maps for 2D-CNN, 3D-CNN, SSRN, HybridSN and SE-HybridSN respectively.

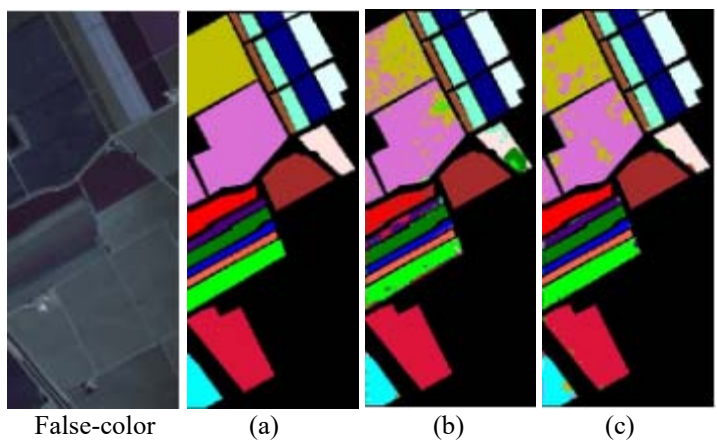

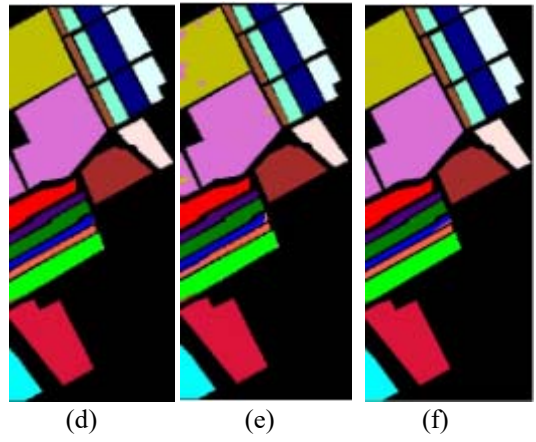

(d)           (e)           (f)

**Fig. 8.** The classification maps of Salinas. (**a**) Ground truth. (**b–f**) Predicted classification maps for 2D-CNN, 3D-CNN, SSRN, HybridSN and SE-HybridSN respectively.

**Table 4.** Classification results in Indian Pines

| Number | Training Samples | 2D-CNN | 3D-CNN | SSRN | HybridSN | SE-HybridSN |
|---|---|---|---|---|---|---|
| 1 | 3 | 12.07 | 27.07 | 79.23 | 61.10 | 45.73 |
| 2 | 72 | 78.46 | 83.45 | 88.60 | 92.20 | 95.44 |
| 3 | 41 | 60.00 | 75.37 | 85.81 | 96.48 | 97.41 |
| 4 | 12 | 42.84 | 56.06 | 70.53 | 77.11 | 93.17 |
| 5 | 24 | 81.87 | 92.90 | 93.11 | 94.30 | 96.71 |
| 6 | 37 | 92.30 | 96.50 | 96.43 | 97.27 | 99.30 |
| 7 | 2 | 27.40 | 67.80 | 82.36 | 89.00 | 98.60 |
| 8 | 24 | 99.44 | 98.27 | 98.51 | 97.97 | 99.00 |
| 9 | 1 | 53.61 | 60.28 | 68.90 | 83.89 | 64.44 |
| 10 | 48 | 74.42 | 83.22 | 87.72 | 95.18 | 96.01 |
| 11 | 122 | 82.74 | 89.38 | 91.42 | 97.78 | 98.31 |
| 12 | 30 | 57.36 | 63.55 | 90.04 | 86.25 | 91.95 |
| 13 | 10 | 84.19 | 88.43 | 91.00 | 89.00 | 98.70 |
| 14 | 63 | 92.57 | 97.89 | 97.96 | 98.23 | 99.43 |
| 15 | 20 | 64.65 | 81.57 | 82.57 | 83.04 | 90.94 |
| 16 | 4 | 81.85 | 92.98 | 88.51 | 85.42 | 96.13 |
| | KAPPA | 0.754±0.03 | 0.84±0.025 | 0.89±0.012 | 0.935±0.008 | 0.963±0.005 |
| | OA (%) | 78.48±3.58 | 86.04±2.19 | 93.10±0.42 | 94.31±0.65 | 96.76±0.44 |
| | AA (%) | 64.74±3.21 | 78.42±2.87 | 88.58±0.15 | 89.01±1.23 | 91.39±2.09 |

**Table 5.** Classification results in the University of Pavia

| Number | Training Samples | 2D-CNN | 3D-CNN | SSRN | HybridSN | SE-HybridSN |
|--------|------------------|--------|--------|------|----------|-------------|
| 1 | 66 | 91.32 | 92.83 | 94.72 | 91.78 | 96.79 |
| 2 | 186 | 97.50 | 96.54 | 97.15 | 99.77 | 99.74 |
| 3 | 21 | 68.51 | 70.08 | 82.73 | 92.24 | 91.44 |
| 4 | 30 | 95.09 | 95.99 | 96.82 | 91.11 | 94.18 |
| 5 | 14 | 99.19 | 99.72 | 99.71 | 97.76 | 99.82 |
| 6 | 50 | 89.59 | 80.84 | 90.48 | 99.38 | 99.31 |
| 7 | 13 | 62.90 | 69.64 | 87.73 | 96.83 | 95.52 |
| 8 | 37 | 87.04 | 80.31 | 88.29 | 90.72 | 93.55 |
| 9 | 10 | 97.75 | 96.70 | 99.90 | 92.17 | 93.86 |
| | KAPPA | 0.854±0.017 | 0.84±0.015 | 0.925±0.017 | 0.946±0.013 | 0.969±0.005 |
| | OA (%) | 89.01±1.48 | 90.04±1.39 | 93.94±0.18 | 95.41±0.95 | 97.55±0.24 |
| | AA (%) | 79.84±2.21 | 88.63±1.85 | 91.98±1.41 | 92.94±2.13 | 96.05±0.78 |

**Table 6.** Classification results in Salinas

| Number | Training Samples | 2D-CNN | 3D-CNN | SSRN | HybridSN | SE-HybridSN |
|--------|------------------|--------|--------|------|----------|-------------|
| 1 | 20 | 59.97 | 97.63 | 97.50 | 99.99 | 99.99 |
| 2 | 37 | 99.48 | 99.82 | 100 | 100.00 | 100 |
| 3 | 20 | 60.01 | 92.35 | 99.23 | 99.48 | 99.54 |
| 4 | 14 | 98.27 | 98.87 | 98.00 | 98.11 | 99.13 |
| 5 | 27 | 94.80 | 96.85 | 97.11 | 99.08 | 98.92 |
| 6 | 40 | 99.89 | 99.98 | 98.43 | 99.57 | 99.93 |
| 7 | 36 | 97.21 | 98.80 | 97.36 | 99.46 | 99.70 |
| 8 | 112 | 83.53 | 87.19 | 98.51 | 99.13 | 98.44 |
| 9 | 62 | 99.26 | 99.55 | 99.90 | 99.97 | 100 |
| 10 | 33 | 84.95 | 93.58 | 97.72 | 98.70 | 97.84 |
| 11 | 11 | 90.00 | 91.44 | 97.42 | 98.34 | 99.04 |
| 12 | 19 | 97.15 | 99.26 | 98.04 | 99.68 | 99.89 |
| 13 | 9 | 95.74 | 97.74 | 97.70 | 97.21 | 94.93 |
| 14 | 10 | 94.84 | 98.29 | 98.42 | 97.58 | 93.51 |
| 15 | 72 | 72.57 | 78.62 | 97.18 | 98.45 | 96.84 |
| 16 | 18 | 91.12 | 87.12 | 91.11 | 99.85 | 99.45 |
| | KAPPA | 0.862±0.02 | 0.918±0.015 | 0.942±0.005 | 0.992±0.003 | 0.986±0.003 |
| | OA (%) | 87.61±1.38 | 92.58±1.19 | 96.10±0.50 | 99.25±0.65 | 98.76±0.24 |
| | AA (%) | 88.66±2.21 | 94.82±1.87 | 97.01±0.63 | 99.09±0.49 | 98.57±0.42 |

We further compared the experiment results of the three 3D-2D-CNN based models and drew the following conclusions. Firstly, the classification accuracy of SE-HybridSN was relatively balanced on three datasets. It further demonstrated the strong feature extraction ability of the dense block and the necessity of Channel Attention module. What is more, SE-HybridSN had an uneven classification accuracy on different datasets. Using a similar amount of training samples in three datasets, the classification effect of Salinas was far better than Indian Pines. Thus, the generalization ability of SE-HybridSN needs to be further analyzed. Thirdly, compared with the other two 3D-2D-CNN models, SE-HybridSN had a tremendous improvement in small sample classes, such as the Stone-steel Towers in Indian Pines and Shadows in the University of Pavia. However, the classification accuracy of SE-HybridSN on some ground objects, such as oats and alfalfa in Indian Pines and Lettuce_romaine_7wk in Salinas was still lower than HybridSN, which needs to be further studied.

## 4    Conclusion

In this paper, in order to realize the efficient extraction and refinement of the spatial–spectral feature in the "small sample" hyperspectral image classification, we proposed an SE-HybridSN model from the perspective of network optimization. Based on the 3D-2D-CNN model, multifeature reuse was realized by a dense block. Besides, the 3D convolution and 2D convolution were respectively equipped with channel attention, thus the spatial–spectral features were further refined. We conducted a series of experiments on three open datasets: Indian Pines, Salinas and the University of Pavia. The experiment results show that the SE-HybridSN model had a better classification effect than all the contrast models. However, the accuracy improvement brought by network optimization was limited, so other strategies should be combined to further improve the classification accuracy.

## References

1. Krizhevsky, A.; Sutskever, I.; Hinton, G.E. Imagenet classification with deep convolutional neural networks. In Proceedings of the 25th International Conference on Neural Information Processing Systems, Lake Tahoe, NV, USA, 3–6 December 2012; pp. 1097–1105.
2. Li, X.; Shen, X.; Zhou, Y .; Wang, X.; Li, T.-Q. Classification of breast cancer histopathological images using interleaved DenseNet with SENet (IDSNet). PLoS ONE 2020, 15, e0232127.
3. Chen, Y .S.; Jiang, H.L.; Li, C.Y .; Jia, X.P .; Ghamisi, P . Deep feature extraction and classification of hyperspectral images based on convolutional neural networks. IEEE T rans. Geosci. Remote Sens. 2016, 54, 6232–6251.
4. Lee, H.; Kwon, H. Going Deeper with Contextual CNN for Hyperspectral Image Classification. IEEE Trans. Image Process. 2017, 26, 4843–4855. [CrossRef] [PubMed]
5. Zhong, Z.; Li, J.; Luo, Z.; Chapman, M. Spectral-spatial residual network for hyperspectral image classification: A 3-D deep learning framework. IEEE Trans. Geosci. Remote Sens. 2018, 56, 847–858.

6.  Wang, W.; Dou, S.; Jiang, Z.; Sun, L. A Fast Dense Spectral-Spatial Convolution Network Framework for Hyperspectral Images Classification. Remote Sens. 2018, 10, 1608. [C

7.  François, C. Deep Learning with Python, 1st ed.; Posts and Telecom Press: Beijing, China, 2018; p. 6.

8.  Liao, W.; Pizurica, A.; Scheunders, P.; Philips, W.; Pi, Y. Semisupervised Local Discriminant Analysis for Feature Extraction in Hyperspectral Images. IEEE Trans. Geosci. Remote Sens. 2013, 51, 184–198.

9.  Prasad, S.; Bruce, L.M. Limitations of Principal Components Analysis for Hyperspectral Target Recognition. IEEE Geosci. Remote Sens. Lett. 2008, 5, 625–629.

10. Li, W.; Prasad, S.; Fowler, J.E.; Bruce, L.M. Locality-Preserving Dimensionality Reduction and Classification for Hyperspectral Image Analysis. IEEE Trans. Geosci. Remote Sens. 2012, 50, 1185–1198.

11. Samaniego, L.; Bardossy, A.; Schulz, K. Supervised classification of remotely sensed imagery using a modified k-NN technique. IEEE Trans. Geosci. Remote Sens. 2008, 46, 2112–2125.

12. Kumar, S.; Ghosh, J.; Crawford, M.M. Best-bases feature extraction algorithms for classification of hyperspectral data. IEEE Trans. Geosci. Remote Sens. 2001, 39, 1368–1379.

13. Foody, G.M.; Mathur, A. A relative evaluation of multiclass image classification by support vector machines. IEEE Trans. Geosci. Remote Sens. 2004, 42, 1335–1343.

14. Hu, W.; Huang, Y.; Wei, L.; Zhang, F.; Li, H. Deep Convolutional Neural Networks for Hyperspectral Image Classification. J. Sens. 2015, 2015, 1–12.

15. Zhao, W.; Guo, Z.; Yue, J.; Zhang, X.; Luo, L. On combining multiscale deep learning features for the classification of hyperspectral remote sensing imagery. Int. J. Remote Sens. 2015, 36, 3368–3379.

16. Liu, B.; Yu, X.; Zhang, P.; Tan, X. Deep 3D convolutional network combined with spatial-spectral features for hyperspectral image classification. Acta Geod. Cartogr. Sin. 2019, 48, 53–63.

17. Meng, Z.; Li, L.; Jiao, L.; Feng, Z.; Tang, X.; Liang, M. Fully Dense Multiscale Fusion Network for Hyperspectral Image Classification. Remote Sens. 2019, 11, 2718. [CrossRef]

18. Swalpa, K.R.; Gopal, K.; Shiv, R.D.; Bidyut, B.C. HybridSN: Exploring 3-D-2-D CNN Feature Hierarchy for Hyperspectral Image Classification. IEEE Geosci. Remote Sens. Lett. 2019, 17, 277–281.

19. Feng, F.; Wang, S.; Wang, C.; Zhang, J. Learning Deep Hierarchical Spatial–Spectral Features for Hyperspectral Image Classification Based on Residual 3D-2D CNN. Sensors 2019, 19, 5276.

20. Hu, J.; Shen, L.; Sun, G. Squeeze-and-excitation networks. In Proceedings of the IEEE Conference on Computer Vision and Pattern Recognition, Salt Lake City, UT, USA, 18–22 June 2018; pp. 7132–7141.

21. Woo, S.; Park, J.; Lee, J.Y.; Lee Kweon, I. Cbam: Convolutional block attention module. In Proceedings of the European Conference on Computer Vision (ECCV), Munich, Germany, 8–14 September 2018; pp. 3–19.

22. Wang, L.; Peng, J.; Sun, W. Spatial–Spectral Squeeze-and-Excitation Residual Network for Hyperspectral Image Classification. Remote Sens. 2019, 11, 884.

23. Li, G.; Zhang, C.; Lei, R.; Zhang, X.; Ye, Z.; Li, X. Hyperspectral remote sensing image classification using three-dimensional-squeeze-and-excitation-DenseNet (3D-SE-DenseNet). Remote Sens. Lett. 2020, 11, 195–203.

24. Lin, Z.; Chen, Y.; Ghamisi, P.; Benediktsson, J.A. Generative Adversarial Networks for Hyperspectral Image Classification. IEEE Trans. Geosci. Remote Sens. 2018, 56, 5046–5063.
25. Zhong, Z.; Li, J.; Clausi, D.A.; Wong, A. Generative Adversarial Networks and Conditional Random Fields for Hyperspectral Image Classification. IEEE T. Cybern. 2020, 50, 3318–3329.
26. Liu, B.; Yu, X.; Zhang, P.; Tan, X.; Yu, A.; Xue, Z. A semi-supervised convolutional neural network for hyperspectral image classification. Remote Sens. Lett. 2017, 8, 839–848.
27. Sellami, A.; Farah, M.; Farah, I.R.; Solaiman, B. Hyperspectral imagery classification based on semi-supervised 3-D deep neural network and adaptive band selection. Expert Syst. Appl. 2019, 129, 246–259.
28. Song,W.; Li, S.; Li, Y. Hyperspectral images classification with hybrid deep residual network. In Proceedings of the 2017 IEEE International Geoscience and Remote Sensing Symposium (IGARSS), Fort Worth, TX, USA, 23–28 July 2017; pp. 2235–2238
29. Liu, B.; Yu, X.; Yu, A.; Zhang, P.Q.; Wan, G.; Wang, R.R. Deep Few-Shot Learning for Hyperspectral Image Classification. IEEE Trans. Geosci. Remote Sens. 2019, 99, 1–15.